\documentclass[graybox]{svmult}

\usepackage{type1cm}        
\usepackage{makeidx}         
\usepackage{graphicx}        
\usepackage{multicol}        
\usepackage{multirow}
\usepackage[bottom]{footmisc}

\usepackage{newtxtext}       %
\usepackage[varvw]{newtxmath}       
\usepackage[numbers]{natbib}
\usepackage{url}

\makeindex             

\begin{document}

\title*{Improving Automatic Speech Recognition for Non-Native English with Transfer Learning and Language Model Decoding}
\titlerunning{Automatic Speech Recognition for Non-Native English}
\author{Peter Sullivan, Toshiko Shibano, Muhammad Abdul-Mageed}
\institute{Peter Sullivan \at The University of British Columbia, BC, Canada \email{prsull@student.ubc.ca}
\and Toshiko Shibano \at  The University of British Columbia, BC, Canada \email{tshibano@student.ubc.ca}
\and Muhammad Abdul-Mageed \at  The University of British Columbia, BC, Canada \email{muhammad.mageed@ubc.ca}
}
%
%
\maketitle

\abstract{ASR systems designed for native English (L1) usually underperform on non-native English (L2). To address this performance gap, \textbf{(i)} we extend our previous work to investigate fine-tuning of a pre-trained wav2vec 2.0 model~\cite{baevski2020wav2vec,xu2021self} under a rich set of L1 and L2 training conditions. We further \textbf{(ii)} incorporate language model decoding in the ASR system, along with the fine-tuning method. Quantifying gains acquired from each of these two approaches separately and an error analysis allows us to identify different sources of improvement within our models. We find that while the large self-trained wav2vec 2.0 may be internalizing sufficient decoding knowledge for clean L1 speech~\cite{xu2021self}, this does not hold for L2 speech and accounts for the utility of employing language model decoding on L2 data.}


\section{Introduction}
\label{sec:intro}
Although non-native English speakers (L2) outnumber native English speakers (L1) ~\cite{crystal2003english}, major challenges contribute to a gap between performance of automatic speech recognition (ASR) systems on L2 speech. This is mainly due to influence of L1 pronunciation on the learned language, and lack of annotated L2 speech data on which ASR systems can be trained~\cite{radzikowski2021accent, viglino2019end}. To meet these challenges, previous work has generally followed two distinct approaches. The first is to make L2 speech representations more closely match those of L1 speech~\cite{radzikowski2021accent}. The second approach leverages L2 speech data to improve model robustness. Due to L2 data scarcity, this second approach necessitates employment of transfer learning or domain adaptation~\cite{shi2021accented,sun2018domain}.


\begin{figure}[b]
\begin{center}
\includegraphics[width=8cm]{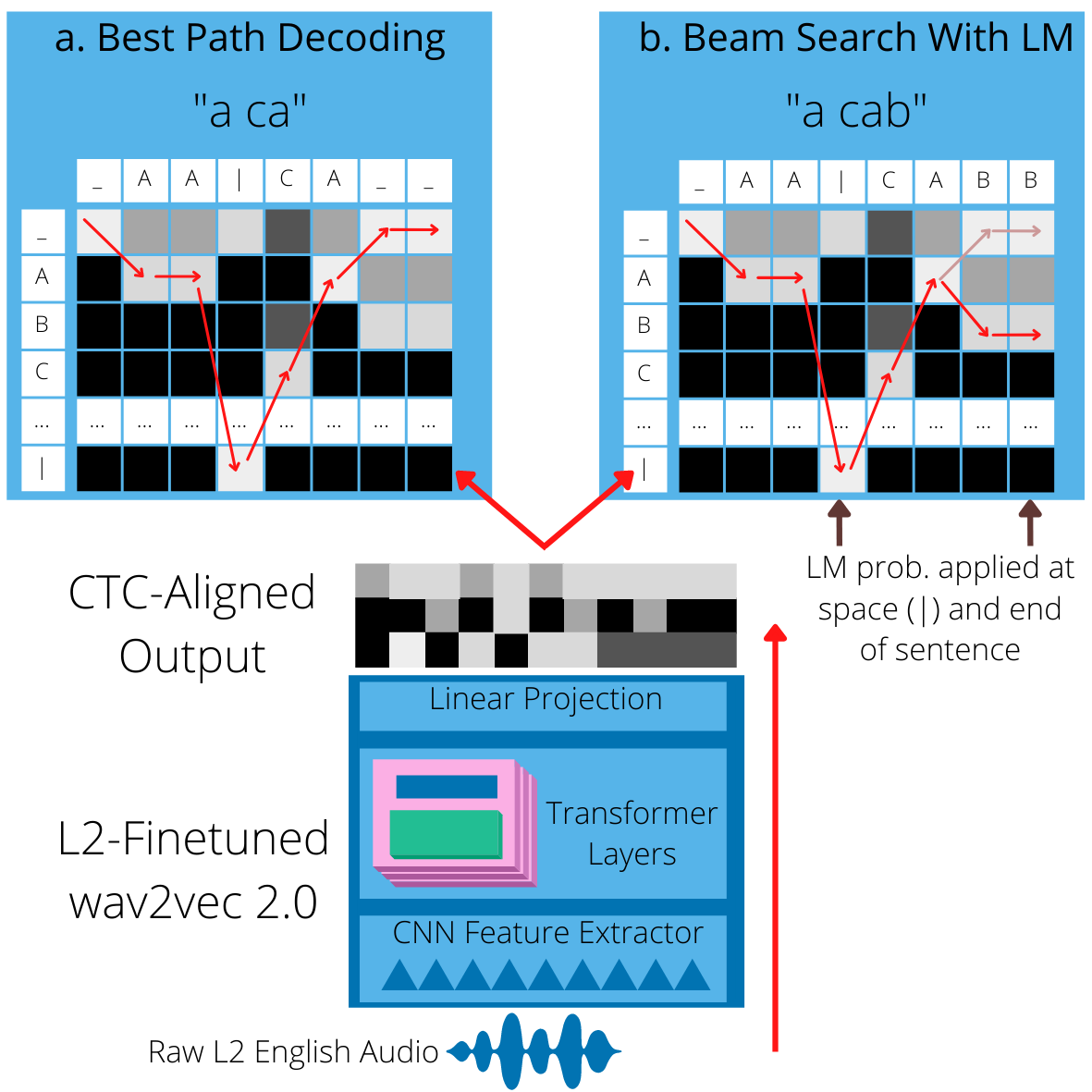}
\caption{\label{fig:overview}
The overall ASR pipeline. We \textbf{(a)} evaluate performance of wav2vec 2.0 without LM using \textit{best path} decoding. We also \textbf{(b)} incorporate language model decoding with \textit{beam search} along with the fine-tuned model.
}
\end{center}
\end{figure}


State-of-the-art ASR models based on self-supervised pre-training such as wav2vec ~\cite{schneider2019wav2vec} and wav2vec 2.0~\cite{baevski2020wav2vec}\footnote{Although sometimes referred to as `unsupervised', these models employ a self-supervised objective.} offer a tantalizing starting point for applying the transfer learning approach we list above, especially due to their strong performance of self-trained wav2vec 2.0 models on ASR in low-resource settings even without a language model~\cite{xu2021self}. However, challenges remain in identifying how best to apply models such as wav2vec 2.0 in L2 fine-tuning scenarios. In spite of this advantage of a fine-tuned model, it is not clear whether the knowledge it acquires is orthogonal to that of a language model especially on L2 speech. Hence, we are interested in \textit{investigating the practical sufficiency of fine-tuned models on their own, and the extent to which they may benefit from external language model decoding on both L1 and L2 speech}. As such, our main objective in the current work is to investigate a rich set of conditions under which we can fine-tune ASR models for optimal L2 performance and the utility of integrating language model decoding along with fine-tuning in an overall ASR model. Concretely, we pursue this primary objective through the following sub-objectives:

\begin{enumerate}
    \item Evaluate fine-tuning and language model decoding strategies for adapting pre-trained L1 English ASR models to L2 English;
    \item Explore the impact of non-native (L2)  accents on performance of these ASR models fine-tuned under various conditions, comparing \textit{multi-accent} training to \textit{single-accent} training; 
    \item Quantify the impact of L2 fine-tuning on model performance for L1 English speech recognition; and
    \item Analyze error categories associated with fine-tuning, as well as language model-decoding.
\end{enumerate}

 Our investigation of the role of language-model decoding in L2 ASR performance extends our previous work~\cite{shibano2021speech}. We also better contextualize the magnitude of impact of fine-tuned only vs. fine-tuning+LM decoding on the downstream tasks for both L1 and L2 speech. The rest of the paper is organized as follows: Section~\ref{sec:lit} is an overview of related work. We introduce our methods in Section~\ref{sec:methods}. We describe our data in Section~\ref{sec:data}, and Section~\ref{sec:exp} is about our experiments and results. We conclude in Section~\ref{sec:con}. 

\section{Related Work}
\label{sec:lit}

Because of the difficulty in linguistically annotating corpora for Hidden Markov Model (HMM)-based ASR ~\cite{graves2014towards}, researchers have broadly embraced End-to-End (E2E) deep learning architectures either based on Connectionist Temporal Classification (CTC) ~\cite{graves2006connectionist,graves2014towards}, Attention ~\cite{chorowski2015attention,chan2016listen, gulati2020conformer}, or hybrids of the two ~\cite{watanabe2017hybrid,wang2020transformer}. Recent efforts inspired by work such as BERT ~\cite{devlin2018bert} have improved on these purely supervised learning baselines through self-supervised pre-training ~\cite{schneider2019wav2vec,baevski2019vq, baevski2020wav2vec} and self-training ~\cite{xu2021self}. These self-supervised wav2vec models represent one line of research in speech representation. Other works include models similar to wav2vec that also use a contrastive loss~\cite{oord2018representation}, models using an autoregressive loss function~\cite{ling2020deep,chung2019unsupervised}, as well as models using a masked language model closer to the original BERT ~\cite{liu2020mockingjay}.

With these efforts, ASR technologies for native languages have evolved significantly. However, we still observe problems in many applications. In particular, several researchers have emphasized how performance of ASR models drops when the input speech is from non-native speakers whose native languages are different from the models' target languages~\cite{radzikowski2021accent,livescu2000lexical,wang2003comparison,ping2008automatic,wang2020laix}. For systems developed for English ASR, this can be a real issue due to the large populations of English language speakers who are non-native~\cite{crystal2003english}. In line with this argument,~\citet{ping2008automatic} points out the necessity to improve speech recognition technology for L2 speakers given that many people speak more than one language for economic and social reasons. It is hoped that continued efforts aiming at improving ASR for non-native speakers will eventually lead to improved results for many as voice recognition technology becomes increasingly pervasive in our daily lives~\cite{ping2008automatic}.

There are two distinct approaches to improve current ASR performance on L2 speech: \textbf{(i)} accent conversion as an extension to the active area of research of voice conversion; and \textbf{(ii)} incorporation of L2 speech data, which is often limited in quantity and quality, during the model training process. The first approach takes inspiration from voice conversion, but instead of focusing on modifying the pitch, it modifies the pronunciation to reduce accents. Additionally, voice conversion models aim to generate results that are speaker-dependent, while accent conversion models deal with generalizing accents from a group of speakers, hence being speaker-independent. With this approach, the resulting model can be used as a pre-processing step to remove accents in the data prior to feeding these data into an ASR model.~\citet{bearman2017accent} adopt this approach but focus on L1 English accents, while ~\citet{radzikowski2021accent} work on L2 English accents with speakers' L1 being Japanese. ~\citet{liu2020end} took a step further and turned Hindi-accented English to native American English without utilizing native utterances.

The second approach often employs methods such as domain adversarial training and transfer learning in order to utilize as much available accented speech data as possible. Domain adversarial training (DAT) is a popular approach as it encourages models to learn accent-invariant features~\cite{sun2018domain, hou2019domain, hu2021redat}. Transfer learning is another popular approach in L2 speech recognition, as it possibly allows a model to gain knowledge from both the base task and the new task, even when the new task has limited data~\cite{matassoni2018non, das2021best, shi2021accented}. In the Accented English Speech Recognition Challenge 2020 (AESRC2020), many teams utilize transfer learning to tackle the L2 accent recognition task~\cite{shi2021accented}. In a recent work, ~\citet{das2021best} combine both DAT and transfer learning to achieve robust accented speech recognition performance.

One method that is common in ASR systems is language model decoding, which re-weights output probabilities to account for greater likelihoods of words occurring in the language. Language models such as KenLM~\cite{heafield2011kenlm}, give probabilities of tokens occurring in a sequence, and thus represent corpus-level statistic of language. Language model decoding can help prevent unlikely sequences of words from being selected (\textit{"the mat chased the rat"}) in favor of more likely predictions (\textit{"the cat chased the rat"}).

While integration of language models has been a standard part of ASR systems, only recent works have been able to reach parity without using an explicit language model, either through knowledge distillation techniques ~\cite{futami2020distilling}, data augmentation ~\cite{park2019specaugment}, or self-training ~\cite{xu2021self, synnaeve2019end}. Language model-free ASR systems are appealing due to the simplicity, but most still struggle with difficult ASR tasks, such as the noisy recordings of LibriSpeech \textit{dev/test-other}. To our knowledge there has been no work to date examining whether the properties of these systems transfers to L2 ASR.

\section{Methods}
\label{sec:methods}
We provide a background about our main methods in this section. We first introduce transfer learning for ASR, then follow with CTC and language model decoding. 
\subsection{Transfer Learning}
For tasks with limited labeled data, training models from scratch becomes impractical. One approach that has great success is transfer learning. Transfer learning involves taking an existing model trained on one or more tasks from a given domain and transferring its knowledge to a target downstream task or domain~\cite{pan2009survey}. Tasks which share the same label and feature space, but perhaps differ in feature distribution, can allow for a simple transfer learning method called model adaptation~\cite{wang2015transfer}. This allows for simply taking an existing model and re-training (i.e., 'fine-tuning') it using a smaller domain-specific dataset. Model adaptation for ASR can be performed easily by freezing part of an existing model and re-training the rest on the new domain~\cite{kunze2017transfer}.

One particularly promising base model for transfer learning is wav2vec 2.0~\cite{baevski2020wav2vec}, which is composed of a multi-layer convolutional neural network feature extractor and a Transformer context network. The network uses a contrastive task for self-supervised pre-training to learn good general representations of audio. Following pre-training, the CNN feature extractor layers of the model are frozen, and the model is fine-tuned on domain specific tasks by adding a linear layer on top of the Transformer context network followed by training with CTC loss~\cite{baevski2020wav2vec}. 

While the original models are strong baselines, the self-trained wav2vec 2.0 Large (LV-60) version of the model~\cite{xu2021self}, which we will refer to as \textit{Wav2Vec 2.0-ST}~\cite{xu2021self}\footnote{\url{https://github.com/pytorch/fairseq/tree/master/examples/wav2vec}}, extends the original work with wav2vec 2.0 by applying a self-training approach. The model is pre-trained on $960$ hours of speech data from LibriSpeech~\cite{panayotov2015librispeech}, followed with self-training on $53.2$k hours of Libri-Light~\cite{kahn2020libri}. During the self-training process pseudo-labels are generated using language models trained on the LibriSpeech corpus, allowing for transfer of knowledge from the language model into the ASR model proper ultimately resulting in a model with little need for an external model during inference time ~\cite{xu2021self}. 

Fine-tuning of pre-trained wav2vec 2.0 is performed with CTC and the transcriptions of the audio segments. For each model, we identify the optimal hyperparameters on the respective Dev set. We choose hyperparameters as follows: For \texttt{mask\_feature\_prob}, we pick from \textit{\{0.25, 0.5\}}, for \texttt{mask\_feature\_length}, we choose from \textit{\{15, 30\}}, for \texttt{mask\_time\_prob} we use \textit{\{0.5, 0.75\}}, and a batch size of $16$. To mimic the tri-state learning rate schedule~\cite{baevski2020wav2vec}, we set different learning rates for different stages:  warm-up (1e-5, 3e-5), constant stage (1e-5, 3e-5), and decay (1e-5, 3e-5, 5e-6). The decay stage is followed by another constant stage (1e-5, 2e-6, 5e-6) to simulate the Fairseq's fine-tuning configuration.

\subsection{CTC Decoding}
Because the output of CTC-trained models is a table of character probabilities for each timestep, this output must be decoded to find the most probable sequence of characters. One simple approach is to use a best path decoding strategy (see top left of Fig. ~\ref{fig:overview}), which simply involves outputting the highest probability token for each timestep condensing duplicate tokens and removing CTC blank symbols ~\cite{graves2012connectionist}. Following~\citet{graves2012connectionist}, we can write the decoding as:
\begin{equation}
    W^*= \text{arg\,}\underset{W}{\text{max}} p(W|X)
\end{equation}
Where $W^*$ is our most likely sequence of characters (the labeling) and $p(W|X)$ is our probability of a labeling given a signal $X$. Then we can write best path decoding as:
\begin{equation}
    W^* \approx \mathcal{F}(\pi^*)
\end{equation}
Where $\mathcal{F}$ is the CTC collapsing function, which removes duplicate letters and \textit{blanks} tokens, and $\pi*$ is the highest activation in the CTC output at a given timestep.
The simplicity of this method allows for fast predictions, but at the cost of potential errors added through not considering combined probability states. As ~\citet{graves2012connectionist} notes, this matters when the "label is weakly predicted for several consecutive timesteps"(p. 71). 
Several algorithms have been introduced to fix this shortcoming: \textit{prefix search}, which allows for accounting for the probability of children nodes in the search graph ~\cite{graves2012connectionist}; \textit{token passing}, which allows integration of a dictionary  ~\cite{graves2012connectionist}, and \textit{decoding with attention}, which uses a secondary RNN model to correct errors ~\cite{zenkel2017comparison}. 

Many decoding strategies aim to also integrate a language model in the process, which allows for incorporating lexical information into the decoding process. N-gram language models can be formalized as: 
\begin{equation}
    p(w_i|w_{i-1},w_{i-2}...,w_{i-n-1})
\end{equation}
Where $w_i$ is the $i$th word in the sequence, which we would like to estimate the probability of, and $n$ is our n-gram size. Probabilities are generally calculated from a text corpus either through efficient stastical packages such as KenLM~\cite{heafield2011kenlm} or through training neural networks to generate probability distributions of the tokens.
Additional decoding strategies that use language model probability re-weighting include modified beam search strategy ~\cite{graves2014towards,hannun2014first}, weighted finite state transducers ~\cite{miao2015eesen, sak2015learning}, character-level recurrent neural network (RNN) decoding ~\cite{maas2015lexicon, hwang2016character}, or word-level RNN decoding ~\cite{hori2018end}.

In our experiments, we choose to apply the prefix beam search strategy for both decoding and including an external language model (see top right of Fig. ~\ref{fig:overview}). Instead of rehashing the full prefix beam search algorithm (see ~\cite{hannun2014first}), we focus on the main components needed to understand the hyperparameter optimization process of this decoding strategy. Prefix beam search attempts to find transcriptions which maximize the following equation (see~\cite{hannun2014first}):
\begin{equation}
    p_{CTC}(W;X)p_{LM}(W)^\alpha|W|^\beta
    \label{eq:prefix_decoding}
\end{equation}
Here $p_{CTC}(W;X)$ is our CTC-trained neural network probability of a character sequence $W$ given an input audio $X$, $p_{LM}(W)$ is the language model probability of sequence $W$, $\alpha$ is the language model weight term, and $\beta$ is a word insertion penalty modifier. The algorithm to maximize the value in~\ref{eq:prefix_decoding}, is similar to normal beam search in the sense that it keeps track of a set of possible contenders $\leq k$, where we call $k$ the beam width ~\cite{lowerre1976harpy,meister2020if}. For CTC, the complexities of duplicates and \textit{blank} tokens mean that the actual probability of a given proposed sequence needs to be calculated as follows:
\begin{equation}
    p(l; x_{1:t}) = (p_b(l; x_{1:t})+p_{nb}(l; x_{1:t}))|W(l)|^\beta
    \label{eq:prefix_probs}
\end{equation}
Where $p(l; x_{1:t})$ is the probability of a given prefix, $p_b(l;x_{1:t})$ is the probability of a $blank$ token being appended onto the current sequence, $p_{nb}(l;x_{1:t})$ is the probability of the next token being a character or punctuation (i.e. $non-blank$), and $|W(l)|^\beta$ is our word insertion term based on the words $W(l)$ in our proposed sequence $l$. A list of these probabilities is kept and updated based on the probabilities of each of the characters in the next segment of the CTC output table. When \textit{space} characters are added to an existing sequence, the language model probability weight $p(W(l^+)|W(l))^\alpha$ is multiplied to the probability of the sequence, where $W(l^+)$ is the new set of words in the sequence. Values of $\alpha$, which indicates how much to emphasize the language model, and $\beta$, which must be set via a hyperparameter tuning process.

In our experiments with adding language model decoding, we use the \textit{pyctcdecode}\footnote{\url{https://github.com/kensho-technologies/pyctcdecode}} implementation of prefix beam search. It functions much the same way as normal prefix beam search, differing only in several minor ways: first by using caching to speed up the decoding process and second by adding a partial word score which penalizes out of vocabulary word matches (based on checking whether the prefix is in a trie of the unigram vocabulary). For hyperparameter tuning, we perform a small grid search using the development set of L2-ARCTIC, with the ranges $\alpha \in \{0.5, 1, 1.5\}$ (considering both downweighting and over-emphasizing the LM), $\beta \in \{0.5, 1, 1.5\}$ and $beamwidth \in \{50, 100, 150, 200\}$, with final hyperparameters as $\alpha = 1, \beta =  1.5, beamwidth = 200$. For experiments on LibriSpeech, we similarly set hyperparameters on the development set (dev-other) and find the combination $\alpha = 0.5, \beta = 0.5, beamwidth = 100$ works best. To ablate the contribution of the language model, we also conduct an experiment on the full splits of L2-ARCTIC with $\alpha = 0$, effectively neutralizing the impact of the language model, keeping the rest of the hyperparameters the same.

\section{Data}
\label{sec:data}

\subsection{Corpus Information}
We choose \textbf{L2-ARCTIC}, a non-native English speech corpus~\cite{zhao2018l2}, for L2 fine-tuning. The recordings are from 24 non-native speakers of English with a total of six different L1s, and each of the L1s consists of two female speakers and two male speakers. The L1s we use for our experiments are Arabic (AR), Hindi (HI), Korean (KO), Mandarin (ZH), Spanish (ES), and Vietnamese (VI). Because L2-ARCTIC is based on the original L1 English corpus, CMU ARCTIC~\cite{kominek2003cmu} (henceforth \textbf{L1-ARCTIC}, for simplicity), we can easily evaluate performance from fine-tuning on same-domain L1 data.

 Each speaker in L2-ARCTIC contributed approximately one hour of phonetically-balanced read speech based on the L1-ARCTIC prompts, which consist of carefully selected sentences ($1,132$ sentence prompts) from Project Gutenberg~\cite{kominek2003cmu}. We note this, as the pretrained wav2vec 2.0 model we use was first pre-trained on LibriSpeech~\footnote{http://www.openslr.org/12/}~\cite{panayotov2015librispeech} and then self-trained on Libri-Light~\footnote{https://github.com/facebookresearch/libri-light}~\cite{kahn2020libri}.  Both corpora rely on audiobooks from the LibriVox project~\footnote{https://librivox.org}, much of which comes from Project Gutenberg~\footnote{http://www.gutenberg.org}. However, because the ARCTIC corpus was selected to create a good phonological balance of sentences and weighted towards fiction~\cite{kominek2003cmu}, there may be domain mismatch between the sets of texts selected between these different corpora, and we aim to measure this with experiments using L1 fine-tuned models. Finally, we ensure there is no overlap in sentences between our L2-ARCTIC dev and test sets and the LibriSpeech training sets.

We also evaluate our fine-tuned models on \textbf{1) LibriSpeech} to compare the fine-tuning with the original performance of \textit{Wav2Vec 2.0-ST}. In addition, we evaluate on \textbf{2) L1-ARCTIC}, identical to our L2-ARCTIC corpus but spoken by four native US English speakers, allowing us to identify any degradation on same-domain L1 speech performance, as well as estimate potential domain mismatch between the LibriSpeech corpus used to train \textit{Wav2Vec 2.0-ST} and ARCTIC. Each of L1-ARCTIC speakers' datasets contain approximately the same number of utterances ($n=\sim1,132*4$) as each of L2-ARCTIC speakers' datasets.\par
For the purpose of our experiments, we define \textit{native (L1) accents} as those represented in the LibriSpeech and L1-ARCTIC, and \textit{non-native (L2) accents} as those represented in L2-ARCTIC.

\begin{table*}[!t]
\caption{Summary of data splits, fine-tuning, and evaluation setups.}

\label{table:model_summary}
\begin{center}
\begin{tabular}{llcclcc}
 &  & \multicolumn{2}{c}{\textbf{Accent dependency}} &  & \multicolumn{2}{c}{\textbf{Speaker dependency}} \\ \cline{3-4} \cline{6-7} 
 &  & \multicolumn{1}{l}{\textbf{Dependent}} & \multicolumn{1}{l}{\textbf{Independent}} &  & \multicolumn{1}{l}{\textbf{Dependent}} & \multicolumn{1}{l}{\textbf{Independent}} \\ \hline
\textbf{Multi-accent} & Model-1 (Split 1) & x &  &  & x &  \\
 & Model-2 (Split 2) & x &  &  &  & x \\
 & Model-3 (Split 3) &  & x &  &  & x \\ \hline
\textbf{Single-accent} & Model-4 (Split 4) & x & x &  & x & x \\
 & Model-5 (Split 5) & x &  &  &  & x \\ 
\end{tabular}
\end{center}
\end{table*}


\begin{figure}[b]
\begin{center}
\includegraphics[width=10cm]{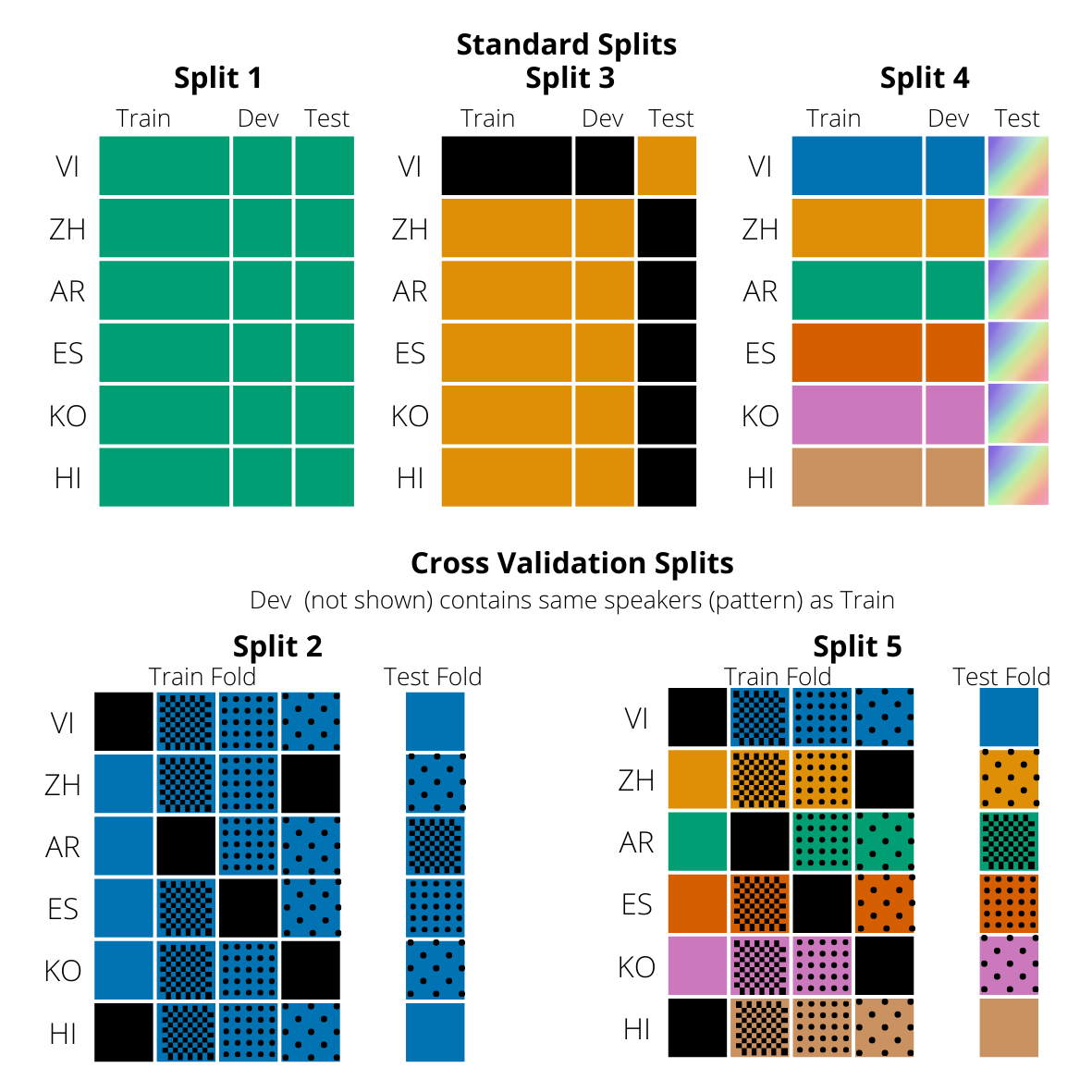}
\end{center}
\caption{\label{fig:splits}
The various data splits we use in our experiments. Color represents a different run of our training, with the rainbow blocks in Split 4 being present in all runs. For cross validation splits, we show a single fold as an example. Speakers are indicated by pattern with 'held-out' speakers blacked out in the training set.
}

\end{figure}
\subsection{Data Splits}
For both L2-ARCTIC and L1-ARCTIC, we split the data into three distinct Train, Dev, and Test sets with an $80$:$10$:$10$ ratio. Importantly, we ensure there is \textit{no overlap between utterances}. For L2-ARCTIC, we split the data across the following settings (see Fig. \ref{fig:splits}).

\begin{itemize}
    \item  \textbf{Split-1} \textit{(speaker-dependent, multi-accent split)}: All speakers from all accents in the Train set are also included in the Dev and Test sets; however, no utterances are shared between Train, Dev, and Test.
    
    \item  \textbf{Split-2} \textit{(speaker-independent cross-validation splits with multiple accents)}: A speaker from each accent\footnote{We use the term `accent' here to loosely refer to variation in speakers with L1 other than English.} is removed from the Train and Dev sets, but other speakers with the same accent remain in the Train and Dev sets. 

    \item  \textbf{Split-3} \textit{(speaker-independent zero-shot splits with multiple accents)}: All speakers from one of the accents are entirely removed from the Train and Dev sets. The removed speakers are included in Test. 
 
    \item   \textbf{Split-4} \textit{(all-speaker, single-accent split)}: Speakers are broken down by accents (six accents in total) and all speakers in a given accent are split into the Train, Dev, and Test sets (3 data splits x 6 accents).
 
    \item   \textbf{Split-5} \textit{(speaker-independent cross-validation splits with single accent)}: One speaker in each accent is removed from the Train and Dev sets, but the other speakers with the same accent remain in the Train and Dev sets. As there are four speakers per accent, four splits are created for each accent, which are further split into the Train, Dev, and Test sets (3 data splits x 6 accents x 4 speakers).

\end{itemize}

\section{Experiments}\label{sec:exp}

 For all our wav2vec 2.0 models, we use Fairseq~\footnote{\url{https://github.com/pytorch/fairseq}} fine-tuning default settings as a reference and convert the hyperparameters to align with Huggingface's implementation. We evaluate all our models in terms of word error rate (WER). For L2 fine-tuning we train each model with three random seeds and report the average WER. Our experiment code is available online~\footnote{\url{https://github.com/UBC-NLP/L2ASR}}.

\subsection{Baselines}
We use the following baselines:

\begin{itemize}

\item  \textbf{Baseline-I:} We use \textit{Wav2Vec 2.0-ST} as a baseline, due to its strong performance on L1 English speech. We use the model released via HuggingFace~\footnote{\url{https://huggingface.co/facebook/wav2vec2-large-960h-lv60-self}}.

\item  \textbf{Baseline-II:} This is \textit{Wav2Vec 2.0-ST}, the same as Baseline-I, fine-tuned on L1-ARCTIC described earlier. The purpose of Baseline-II is to measure potential domain shift between LibriSpeech/Libri-Light and ARCTIC, as well as to measuring potential trade-offs from the fine-tuning process itself. 
\end{itemize}

\subsection{Multi-Accent Models}


With our multi-accent models, we examine performance using multiple accents during training. We introduce each of our models here, and present the results acquired with each. We provide a summary of our different data splits and models across accent and speaker dependency categories in Table~\ref{table:model_summary}.

\begin{table*}[!t]
\caption{\label{model-1} 
    Model-1 performance in word error rate (WER) (lower is better) on non-native accents (L2-ARCTIC) and native accents (L1-ARCTIC, LS\textsubscript{dev} and LS\textsubscript{test}). Baseline-I and Baseline-II are reported on the same Dev and Test sets of each corpus for comparison.
    }
\begin{center}
\begin{tabular}{llrrlrrlrrlrr}
\hline
 &  & \multicolumn{2}{c}{\textbf{L2-ARCTIC}} &  & \multicolumn{2}{c}{\textbf{L1-ARCTIC}} &  & \multicolumn{2}{c}{\textbf{LS\textsubscript{dev}}} &  & \multicolumn{2}{c}{\textbf{LS\textsubscript{test}}} \\ \cline{3-4} \cline{6-7} \cline{9-10} \cline{12-13} 
\textbf{Model} &  & \multicolumn{1}{r}{\textbf{Dev}} & \multicolumn{1}{r}{\textbf{Test}} & \multicolumn{1}{r}{} & \multicolumn{1}{r}{\textbf{Dev}} & \multicolumn{1}{r}{\textbf{Test}} & \multicolumn{1}{r}{} & \multicolumn{1}{r}{\textbf{Clean}} & \multicolumn{1}{r}{\textbf{Other}} & \multicolumn{1}{r}{} & \multicolumn{1}{r}{\textbf{Clean}} & \multicolumn{1}{r}{\textbf{Other}} \\ \hline
Baseline-I &  & 13.47 & 12.47 &  & 2.30 & 2.23 &  & \textbf{1.69} & \textbf{3.55} &  & \textbf{1.86} & \textbf{3.89} \\
Baseline-II &  & 17.29 & 15.95 &  & \textbf{1.26} & \textbf{1.30} &  & 2.19 & 5.13 &  & 2.32 & 5.00 \\
Model-1 &  & \textbf{9.78} & \textbf{9.27} &  & 1.94 & 1.86 &  & 2.75 & 5.55 &  & 2.82 & 6.36 \\ \hline
\end{tabular}
\end{center}
\end{table*}

     \textbf{Model-1 (speaker- and accent-dependent):}
    The model is fine-tuned with Split-1 data to identify any speaker-dependent training impact, as well as an upper limit on performance. In addition to evaluating on L2-ARCTIC Test, we evaluate on L1-ARCTIC Test and LibriSpeech in order to observe any changes in model performance on L1 English. 
    
    As Table~\ref{model-1} shows, our Model-1 achieves best performance on both Dev and Test of \textbf{L2-ARCTIC} as compared to our two baselines. On Test, our Model-1 acquires $25.66\%$ improvement over our Baseline-I wav2vec 2.0 system on L2-ARCTIC ($9.27$ WER for our model vs. $12.47$ WER for Baseline-I). This gain is not surprising and simply means that a model with access to L2 data for fine-tuning will improve over models fine-tuned with L1 data (Baseline-II, which is fine-tuned on L1-ARCTIC) or not-fine-tuned at all (Baseline-I). Nor is performance on \textbf{L1-ARCTIC} surprising: a model fine-tuned with native data (Baseline-II) outperforms one fine-tuned with accented data (our Model-1), both of which outperform a model without fine-tuning (Baseline-I). These results, however, show that in absence of L1 data, L2 data can be valuable for improving ASR model performance even on L1. For \textbf{LibriSpeech}, Baseline-I, which is trained on LibriSpeech data, outperforms the two fine-tuned models (our Model-1 and Baseline-II). The reason is that these two latter models are fine-tuned on a domain that is different from LibriSpeech. That is, fine-tuning models on out-of-domain data will, and as we see here does, result in deterioration of performance on in-domain data. We also note that our Model-1's performance on LibriSpeech is worse than that of Baseline-II on both the `Clean' (LS\textsubscript{Clean}, native speech under quite recording environments), and `Other' (LS\textsubscript{Other}, both noisy environment and accented recordings),  Dev and Test splits. This may be because LibriSpeech is mostly comprised of L1 data and the greater variability on our L2-ARCTIC Train set (24 non-native speakers in our Model-1 vs. 4 native speakers in Baseline-II).
    


    \textbf{Model-2 (speaker-independent, accent-dependent):} While Model-1 mimics a situation where we have some training data from speakers that we serve (i.e., test on), this is rarely a realistic scenario. We instead switch to a speaker-independent (but still \textit{accent-dependent}) setting, Split-2. We carry out four-fold cross-validation with the 24 speakers in the data, every time using 18 speakers (three speakers per accent) in Train\footnote{We use 10\% of the utterances from these 18 speakers for development (Dev).} and six speakers in Test (one per accent). We report the average of the four folds/runs, along with standard deviation. 
    
    As Table~\ref{model-2} shows, Model-2 performance is consistent with Model-1. Our Model-2 outperforms the two baselines on both Dev and Test, reaching $9.96$ WER on Test compared to $12.47$ for Baseline-I and $15.96$ for Baseline-II. These results demonstrate that fine-tuning with multiple accents improves the accented ASR system without access to test speaker data. 
    

     \textbf{Model-3 (speaker- and accent-independent):} To evaluate performance on \textit{unseen} accents, we adopt a zero-shot strategy by removing one accent at a time from both Train and Dev sets and evaluating on the Test set of the removed accent, Split-3. To evaluate model performance on each accent, we conduct six runs in total with one accent removed at a time. 
    
    As Table~\ref{model-3} shows, fine-tuning on accented speech benefits unseen accents and speakers (Model-3 setting). All the multi-accent, zero-shot models outperform Baseline-I and Baseline-II, which means each of the six accents benefit from other accents through this process of transfer learning. Our results also show that, in absence of in-accent data, some unseen accents are easier for the model than others. For example, on Test\textsubscript{zeroshot}, Vietnamese (VI) is the most challenging (with $18.81$ WER) and Hindi (HI) is the least challenging (with only $6.67$ WER). 



\begin{table}[!t]
\caption{\label{model-2} 
    Model-2 cross validated performance on L2-ARCTIC Dev and Test sets, alongside Baseline-I and Baseline-II performance on the same cross validation splits. Mean refers to the average WER over the four runs and SD refers to the standard deviation.
}
\begin{center}

\begin{tabular}{llrrlrr}
\hline
 &  & \multicolumn{2}{c}{\textbf{Dev\textsubscript{L2}}} &  & \multicolumn{2}{c}{\textbf{Test\textsubscript{L2}}} \\ \cline{3-4} \cline{6-7} 
\textbf{Model} &  & \multicolumn{1}{r}{\textbf{Mean}} & \multicolumn{1}{r}{\textbf{SD}} &  & \multicolumn{1}{r}{\textbf{Mean}} & \multicolumn{1}{r}{\textbf{SD}} \\ \hline
Baseline-I &  & 13.47 & 0.23 &  & 12.47 & 0.84 \\
Baseline-II &  & 17.29 & 0.41 &  & 15.96 & 1.58 \\
Model-2 &  & \textbf{9.57} & 0.19 &  & \textbf{9.96} & 0.64 \\ \hline
\end{tabular}
\end{center}
\end{table}

\begin{table*}[!t]
 \caption{\label{model-3}
     Model-3 setting, where a different accent is removed each run. Test\textsubscript{all} refers to Test of \textit{all} 24 speakers, and Test\textsubscript{zeroshot} refers to Test of those four speakers who have L1\textsubscript{removed} accent. Baseline-I acquires $12.47$ on Test\textsubscript{all}, while Baseline-II acquires $15.95$ on the same test set (i.e., Test\textsubscript{all}).
 }
 \begin{center}
\begin{tabular}{cccccc|ccc}
\hline
  &  & \textbf{Baseline-I} &  & \textbf{Baseline-II} &  & \multicolumn{3}{c}{\textbf{Model-3}} \\
  \cline{3-3} \cline{5-5} \cline{7-9}
 \textbf{L1\textsubscript{removed}} &  & \textbf{Test\textsubscript{zeroshot}} &  & \textbf{Test\textsubscript{zeroshot}} &  & \textbf{Dev\textsubscript{L2}} & \textbf{Test\textsubscript{zeroshot}} & \textbf{Test\textsubscript{all}} \\
 \hline
 VI &  & 23.30 &  & 28.81 &  & \textbf{7.96} & 18.81 & 9.43\\
 ZH &  & 14.85 &  & 19.32 &  & 9.02 & 12.13 & 9.08\\
 AR &  & 10.95 &  & 14.82 &  & 9.40 & 10.10 & 9.13\\
 ES &  & 10.48 &  & 13.48 &  & 9.38 & 8.89 & \textbf{8.98}\\
 KO &  & 8.18 &  & 10.22 &  & 10.10 & 6.95 & 9.01\\
 HI &  & \textbf{6.93} &  & \textbf{8.93} &  & 10.29 & \textbf{6.67} & 9.11\\
 \hline
\end{tabular}
\end{center}
\end{table*}

\begin{figure}[b]
\sidecaption
\includegraphics[width=7.5cm]{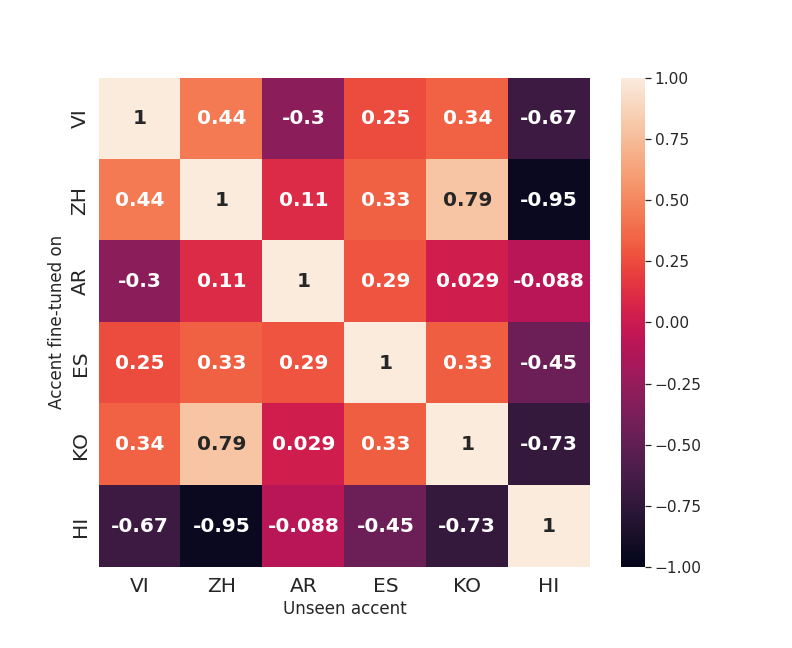}
\caption{Model-4 Correlations of fine-tuning accent vs. test-accent percent performance change. Here we present the correlations based on Table \ref{model-4 zeroshot}, to show the accents that most benefit each other. }
\label{fig:model-4 rocket}
\end{figure}



\subsection{Accent-Specific Models}
We evaluate the accent-dependent performance by fine-tuning our models on a single type of L1-specific accent at a time.

\begin{table*}[!t]
\caption{Model-4 performance on L2 accent (Test\textsubscript{L2}) and native accent (Test\textsubscript{L1}, LS\textsubscript{Clean}, LS\textsubscript{Other}), compared with Baseline-I, Baseline-II, and Model-1. SD refers to the standard deviation.}
\label{model-4}
\begin{center}

\begin{tabular}{llrlrlrlr|rrr}
\hline
 &  & \textbf{Baseline-I} &  & \textbf{Baseline-II} &  & \textbf{Model-1} &  & \multicolumn{4}{c}{\textbf{Model-4}} \\ \cline{3-3} \cline{5-5} \cline{7-7} \cline{9-12} 
\textbf{L1} &  & \textbf{Test\textsubscript{L2}} &  & \textbf{Test\textsubscript{L2}} &  & \textbf{Test\textsubscript{L2}} &  & \textbf{Test\textsubscript{L2}} & \textbf{Test\textsubscript{L1}} & \textbf{LS\textsubscript{Clean}} & \textbf{LS\textsubscript{Other}} \\ \hline
VI &  & 23.30 &  & 28.81 &  & 15.14 &  & \textbf{12.12} & 2.02 & 3.08 & 6.96 \\
ZH &  & 14.85 &  & 19.32 &  & 11.49 &  & \textbf{8.95} & 1.82 & 2.84 & 6.22 \\
AR &  & 10.95 &  & 14.82 &  & 8.90 &  & \textbf{6.92} & 1.55 & 2.66 & 6.24 \\
ES &  & 10.48 &  & 13.48 &  & 8.92 &  & \textbf{6.68} & 1.56 & 2.53 & 6.11 \\
KO &  & 8.18 &  & 10.22 &  & 6.60 &  & \textbf{4.99} & 1.71 & 2.51 & 5.63 \\
HI &  & 6.93 &  & 8.93 &  & 5.51 &  & \textbf{4.99} & 1.52 & 2.36 & 6.05 \\ \hline
\textbf{Mean} &  & 12.45 &  & 15.93 &  & 9.43 &  & 7.44 & 1.70 & 2.66 & 6.20 \\ \hline
\textbf{SD} &  & 5.97 &  & 7.30 &  & 3.49 &  & 2.72 & 0.20 & 0.26 & 0.43 \\ \hline
\end{tabular}
\end{center}
\end{table*}

\begin{table*}[!t]
\caption{Model-4 performance in the zero-shot setting. Bold fonts represent the accent whose WER drops the most in the zero-shot setting. For example, compared with Baseline-I, the VI-specific fine-tuning not only improves performance on VI (i.e., a drop in WER), but also improves on ZH despite ZH being the unseen accent. One notable pattern is that HI-specific fine-tuning only benefits HI-accented speech recognition while all the other fine-tuning hinder performance on the HI accent.}
\label{model-4 zeroshot}
\begin{center}

\begin{tabular}{lrrrrrr}
\hline

 & \multicolumn{1}{c}{\textbf{VI}} & \multicolumn{1}{c}{\textbf{ZH}} & \multicolumn{1}{c}{\textbf{AR}} & \multicolumn{1}{c}{\textbf{ES}} & \multicolumn{1}{c}{\textbf{KO}} & \multicolumn{1}{c}{\textbf{HI}} \\ \hline
\textbf{Baseline-I} & 23.30 & 14.85 & 10.95 & 10.48 & 8.18 & 6.93 \\ \hline
\textbf{VI-specific} & 12.12 & 13.62 & 13.01 & 9.95 & 8.55 & 9.62 \\ 
 \texttt{$\Delta$WER } & -11.18 & -1.23 & 2.06 & -0.53 & 0.37 & 2.69 \\
\texttt{$\Delta \%$ } & -48.00 & \textbf{-8.31} & 18.84 & -5.03 & 4.52 & 38.77 \\ \hline
\textbf{ZH-specific} & 20.37 & 8.95 & 11.42 & 9.79 & 6.82 & 10.91 \\ 
\texttt{$\Delta$WER} & -2.93 & -5.90 & 0.47 & -0.69 & -1.36 & 3.98 \\
\texttt{$\Delta \%$} & -12.58 & -39.75 & 4.26 & -6.62 & \textbf{-16.67} & 57.43 \\ \hline
\textbf{AR-specific} & 23.88 & 14.86 & 6.92 & 9.86 & 9.16 & 7.74 \\ 
 \texttt{$\Delta$WER} & 0.58 & 0.01 & -4.03 & -0.62 & 0.98 & 0.81 \\
\texttt{$\Delta \%$} & 2.47 & 0.07 & -36.83 & \textbf{-5.92} & 11.94 & 11.69 \\ \hline
\textbf{ES-specific} & 20.71 & 13.99 & 11.00 & 6.68 & 7.92 & 8.66 \\ 
\texttt{$\Delta$WER} & -2.59 & -0.86 & 0.05 & -3.80 & -0.26 & 1.73 \\
\texttt{$\Delta \%$} & \textbf{-11.13} & -5.81 & 0.43 & -36.23 & -3.22 & 25.01 \\ \hline
\textbf{KO-specific} & 20.07 & 12.12 & 11.66 & 10.04 & 4.99 & 9.09 \\ 
 \texttt{$\Delta$WER} & -3.23 & -2.73 & 0.71 & -0.44 & -3.19 & 2.16 \\
\texttt{$\Delta \%$} & -13.88 & \textbf{-18.38} & 6.45 & -4.23 & -39.04 & 31.17 \\ \hline
\textbf{HI-specific} & 26.18 & 18.39 & 13.51 & 11.90 & 10.72 & 4.99 \\ 
\texttt{$\Delta$WER} & 2.88 & 3.54 & 2.56 & 1.42 & 2.54 & -1.94 \\
\texttt{$\Delta \%$} & \textbf{12.37} & 23.82 & 23.35 & 13.55 & 31.01 & -27.99 \\ \hline
\end{tabular}
\end{center}
\end{table*}

    \textbf{Model-4 (speaker-dependent, accent-dependent):} The model is fine-tuned with Split-4 data to identify any accent-dependent training impact on downstream performance, as well as an upper bound on performance when the model is optimized for a single accent. In addition to evaluating on L2-ARCTIC Test, we test the model on L1-ARCTIC Test and LibriSpeech as a means to identify any degradation on L1 English data. 
    
    As Table~\ref{model-4} shows, while the multi-accent model (Model-1) outperforms Baseline-I for all six accents, all of the accent-specific models (Model-4 setting) outperform Model-1 on the Test\textsubscript{L2} setting despite the small amount of data (roughly five hours) used for fine-tuning each of the versions of Model-4. On average, Model-4 setting is two points WER better than Model-1. In addition, Model-4 type models (each of which is fine-tuned on one non-native accent) perform reasonably well on L1 data (Test\textsubscript{L1}, LS\textsubscript{Clean}, and LS\textsubscript{Other}). Further, large accent-specific variability is observed across different model types on Test\textsubscript{L2} ($SD$ = [$2.72-7.30$]), compared with native counterparts such as Test\textsubscript{L1} ($SD$ = [$0.20-0.43$]). An interesting result is the apparent difficulty difference between different accents ($HI$ and $KO$ easiest, $VI$ hardest), regardless of model types. We provide sample outputs from Model-4 in Table~\ref{tab:transcripts}. \par
    
    As shown in Table~\ref{model-4 zeroshot}, we also perform accent-wise zero-shot evaluation. Results of this set of experiments reveal an interesting pattern: while fine-tuning on a single accent generally benefits \textit{at least one other accent}, fine-tuning on the Hindi accent only benefits Hindi (the same accent) and hinders performance on \textit{all the other accents}.
    
    Strong to moderate positive correlations (see Fig. \ref{fig:model-4 rocket}) are observed among ZH, KO and VI (0.79 between ZH and KO; 0.44 between VI and ZH; 0.34 between VI and KO). On contrary, HI accents have negative correlations with all the other L1s. Strong negative correlations with ZH, KO and VI (-0.95, -0.73, 0.67, respectively) suggest that the more we fine-tune on HI accents, the more detrimental to the performance on those three accents (and vice versa; those three accents would have negative impacts on HI performance).

 \textbf{Model-5 (speaker-independent and accent-dependent):} This setup simulates a more realistic scenario where we target a single accent, without access to all speakers during development time. Thus, we use Split-5 data which mimics a speaker-independent setting. We cross-validate each L1 subset with one of the four speakers per fold. The hyperparameters we use are those identified for Model-4. To evaluate the performance on each speaker, we conduct $24$ folds in total with one speaker removed at a time, and report the average and standard deviation of the four folds per each accent. As Table~\ref{model-5} shows, speaker-dependent variability is small for Test\textsubscript{all} ($SD$ = [$0.11-0.38$]) but large for Test\textsubscript{zeroshot-speaker} ($SD$ = [$1.12-4.87$]). These results suggest that individual speaker's differences may play an important role in how much performance gain can be obtained by fine-tuning.\footnote{For those speakers whose TOEFL scores are known~\cite{zhao2018l2}, a strong negative correlation was observed between speaker-specific WERs of Baseline-I and speaker's TOEFL scores, $r$($8$) $ \approx -.77$, $p$ \textless $.01$.}

\begin{table}[!t]
\caption{Model-5 performance on L2 accent. Test\textsubscript{all} contains utterances by all speakers within each L1 whereas Test\textsubscript{zeroshot-speaker} contains utterances by a single speaker that is absent in the training phase. Mean refers to the average WER over four folds for each L1, and SD refers to the standard deviation.}
\label{model-5} 
\begin{center}

\begin{tabular}{llrrlrr}
\hline
 &  & \multicolumn{2}{c}{\textbf{Test\textsubscript{all}}} &  & \multicolumn{2}{c}{\textbf{Test\textsubscript{zeroshot-speaker}}} \\ \cline{3-4} \cline{6-7} 
\textbf{L1} &  & \multicolumn{1}{r}{\textbf{Mean}} & \multicolumn{1}{r}{\textbf{SD}} &  & \multicolumn{1}{r}{\textbf{Mean}} & \multicolumn{1}{r}{\textbf{SD}} \\ \hline
VI &  & 12.67 & 0.38 &  & 14.28 & 4.87 \\
ZH &  & 9.65 & 0.31 &  & 11.26 & 3.03 \\
AR &  & 7.28 & 0.29 &  & 8.56 & 2.28 \\
ES &  & 6.95 & 0.26 &  & 7.76 & 3.99 \\
KO &  & 5.22 & 0.18 &  & 5.69 & 2.20 \\
HI &  & 5.27 & 0.11 &  & 5.79 & 1.12 \\ 
\hline
\end{tabular}
\end{center}
\end{table}

\subsection{Language Model Decoding}

\begin{table}[!t]
\caption{Comparison of models with and without language model decoding on the full L2-ARCTIC Test set. We further ablate this by setting $\alpha$ to $0$ to demonstrate performance with beam search, but without language model re-weighting. $\Delta WER$ indicates increase (+) or decrease (-) in WER given as a percent relative to the No LM results}
\label{tab:lm_decode1}
\begin{center}
\begin{tabular}{lll|ll}
\hline 
   & Best Path $\downarrow$ & Beam Search $\downarrow$ & $\Delta WER$   $\downarrow$  \\
   \hline
B1 & 12.47  & 8.43  & -32.40 \% \\
B1$_{\alpha=0}$ & 12.47  & 12.74 & +2.17\% \\
B2 & 15.95  & 11.96 & -25.02 \%  \\
B2$_{\alpha=0}$ & 15.95  & 16.38 & +2.70\% \\

M1 & \textbf{9.27}   & \textbf{5.53} &

\textbf{-40.32\%} \\
M1$_{\alpha=0}$ & 9.27 & 9.42 & +1.62\% \\ \hline 
\end{tabular}
\end{center}
\end{table}

\begin{table}[!t]
\caption{Comparison of models with and without language model decoding on the language background splits (subscript). Results in WER, and relative decrease (-) in WER ($\Delta WER \%$)}
\label{tab:lm_decode2} 
\begin{center}
\begin{tabular}{lll|ll}
\hline 
   & Best Path $\downarrow$ & Beam Search $\downarrow$ & $\Delta WER$ \% $\downarrow$   \\
   \hline
$B1_{VI}$ & 23.30 & 17.01 & -27.00\% \\
$B2_{VI}$ & 28.81 & 22.60 & -21.56\% \\
$M4_{VI}$ & \textbf{12.12} & \textbf{7.36}  & \textbf{-39.23}\% \\
\hline
$B1_{ZH}$ & 14.85 & 9.76 & -34.28\% \\
$B2_{ZH}$ & 19.32 & 14.28 & -26.09\% \\
$M4_{ZH}$ & \textbf{8.95}  & \textbf{5.98}  & \textbf{-33.20\%} \\
\hline
$B1_{AR}$ & 10.95 & 7.53  & -31.23\% \\
$B2_{AR}$ & 14.82 & 10.90 & -26.45\% \\
$M4_{AR}$ & \textbf{6.92}  & \textbf{4.72}  & \textbf{-31.81\%} \\
\hline
$B1_{ES}$ & 10.48 & 7.04  & -32.82\% \\
$B2_{ES}$ & 13.48 & 9.96 & -26.11\% \\
$M4_{ES}$ & \textbf{6.68}  & \textbf{3.96}  & \textbf{-40.80\%} \\
\hline
$B1_{KO}$ & 8.18  & 4.75  & -41.93\% \\
$B2_{KO}$ & 10.22 & 7.25  & -29.06\% \\
$M4_{KO}$ & \textbf{4.99}  & \textbf{2.80}  & \textbf{-43.85\%} \\
\hline
$B1_{HI}$ & 6.93  & 4.43  & -36.08\% \\
$B2_{HI}$ & 8.93  & 6.67  & -25.31\% \\
$M4_{HI}$ & \textbf{4.99}  & \textbf{2.78}  & \textbf{-44.22\%} \\
\hline

\end{tabular}
\end{center}
\end{table}

\begin{figure}[t]
\sidecaption
\includegraphics[width=7.5cm]{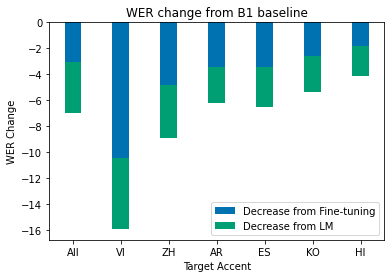}
\caption{\label{fig:wer_change}
Here we show the average absolute WER reduction from adding a language model compared with fine-tuning on the different test splits. 
}
\end{figure}

We evaluate the impact of language model decoding in comparison to the fine-tuning techniques already identified. We use a 4-gram KenLM model ~\cite{heafield2011kenlm} trained on the concatenated LibriSpeech and ARCTIC training corpora. We find performance gain from language model decoding to be relatively similar to fine-tuning for most splits, with the combination of the two methods even more beneficial (see Tables \ref{tab:lm_decode1} \& \ref{tab:lm_decode2}). To further quantify the results, for each target accent, we calculate the average reduction from adding language model decoding and compare with the average reduction from fine-tuning, calculated as $\Delta WER_{LM} = AVG( B1_{LM} - B1 , M_{LM} - M  )$ while fine-tuning reduction is  $\Delta WER_{FT}= AVG(M- B1, M_{LM} -B1_{LM} )$  (see Fig. \ref{fig:wer_change}). For more difficult accents (VI, ZH), fine-tuning appears to play a much bigger role in performance improvements, with easier accents benefiting more from the language model decoding (HI, KO).

In evaluating beam search on its own ($\alpha=0$), performance degrades slightly compared to the baseline. This indicates that most of the performance gain in decoding is coming from the inclusion of the language model. We note the B2 baseline not only performs worse than B1 baseline, but additionally benefits the least from language model decoding (perhaps indicating that it has overfit to the L1-ARCTIC corpus). For L2 ASR, this suggests that simply fine-tuning on domain-specific but L1 accent corpora is counterproductive.

For performance on the LibriSpeech corpus (see Table \ref{tab:lm_decode3}), results are more mixed. As already shown by earlier work~\cite{xu2021self}, \textit{Wav2Vec 2.0-ST} benefits only slightly from language model decoding on the clean split of LibriSpeech ($\Delta WER -0.05$). The fine-tuned models similarly only gain mild benefit, with some ($\Delta WER$ M4ZH: $+0.12$, M4ES: $ +0.02$) models actually performing worse. 
For \textit{test-other}, we observe mild improvements in performance for most models using language model decoding, although some improvements are negligible ($\Delta WER$ M4ZH:  $-0.07$, M4ES: $-0.01$). The overall performance gains from adding a language model and beam search to decoding L1 speech are minor in comparison to the benefits in L2 speech decoding, and fine-tuning on L2 speech decreases the performance on L1 speech substantially even when compared with better decoding (M1 results \textit{test-clean} $40.33$\% and \textit{test-other} $47.68$\% increase from B1). This suggests an \textit{L1 vs. L2 tradeoff} that cannot entirely be overcome by the combination of fine-tuning and decoding strategies we have tried.

\begin{table}[!t]
\caption{Comparison of models with and without language model decoding on the test-clean and test-other splits of LibriSpeech. Results in WER, and relative decrease (-) or increase (+) in WER ($\Delta WER \%$).}
\label{tab:lm_decode3}     
\begin{center}
\begin{tabular}{lrr|r|rr|r}
\hline 
            & \multicolumn{2}{c}{LS Test-Clean} & \multicolumn{4}{c}{LS Test-Other}\\
   &
  \multicolumn{1}{l}{\textbf{Best Path $\downarrow$}} &
  \multicolumn{1}{l}{\textbf{Beam Search $\downarrow$}} &
  \multicolumn{1}{l}{\textbf{$\Delta WER$ \% $\downarrow$ }} &
  \multicolumn{1}{l}{\textbf{Best Path $\downarrow$}} &
  \multicolumn{1}{l}{\textbf{Beam Search $\downarrow$}} &
  \multicolumn{1}{l}{$\Delta WER$ \% $\downarrow$ } \\
  \hline 
B1   & \textbf{1.86} & \textbf{1.81} & -2.69\%  & \textbf{3.89} & \textbf{3.67} & -5.66\%  \\
B2   & 2.32 & 1.96 & \textbf{-15.52}\% & 5.00 & 4.34 & -13.20\% \\
M1   & 2.91 & 2.54 & -12.71\% & 6.44 & 5.42 & \textbf{-15.84}\% \\
M4VI & 2.95 & 2.78 & -5.76\%   & 6.56 & 5.97 & -8.99\%  \\
M4ZH & 2.44 & 2.56 & +4.92\%   & 5.45 & 5.38 & -1.28\%   \\
M4AR & 2.62 & 2.42 & -7.63\%  & 6.13 & 5.38 & -12.23\%  \\
M4ES & 2.29 & 2.31 & +0.87\%   & 5.31 & 5.30 & -0.19\%   \\
M4KO & 2.48 & 2.24 & -9.68\% & 5.53 & 4.82 & -12.84\% \\
M4HI & 2.29 & 2.12 & -7.42\%  & 5.78 & 5.19 & -10.21\% \\ \hline 
\end{tabular}
\end{center}
\end{table}



\begin{table*}[!t]
\small    
\caption{\label{tab:transcripts} 
    Examples of transcription output of selected utterances from the Test set of Model-4 among all six L1s without a language model. Capitalized words indicate errors. We show samples from two speakers per accent.
}\begin{center}
\begin{tabular}{c|l}

\hline
\multicolumn{1}{l|}{\textbf{Model}} & \multicolumn{1}{c}{\textbf{Model output}} \\
\hline
Ref & at lake linderman i had one canoe very good peterborough canoe \\
\hline
\multirow{1}{*}{VI} & at LAY LINDEMAN i had one canoe very good PETERBORROUG CANOES \\
 & A lake LNDER MAN i had one canoe very good BIET OF ROCK canoe \\
\hline
\multirow{1}{*}{ZH} & at lake LINGERMAN i had ONCE canoe very good PETERBROUGH canoe\\
 & at lake LINERMAN i had one canoe very good PETERE BROUGHTA canoe \\
\hline
\multirow{1}{*}{AR} & at lake LUNDERBOGH i had one canoe very good BITTERBOROUGH canoe \\
 & at lake LUNDERMAN i had one canoe very good BETTER BORT canoe \\
\hline
\multirow{1}{*}{ES} & at lake linderman i had one canoe a very good PETERBOURN canoe \\
 & at lake linderman i had ONCE canoe very good PIERREBOROUGH canoe \\
\hline
\multirow{1}{*}{KO} & at lake linderman i had one canoe very good peterborough canoe\\
 & at lake LINDEMAN i had ONCE canoe very good PITTEBRAUG canoe \\
\hline
\multirow{1}{*}{HI} & at lake LINDEMAN i had one canoe very good PETERBURGH canoe \\
 & at lake linderman i had one canoe A very good PEACHERBROROU canoe \\
\hline
\end{tabular}
\end{center}
\end{table*}

\section{Error Analysis}\label{sec:err}
\begin{figure}[t]
\sidecaption
\includegraphics[width=7.5cm]{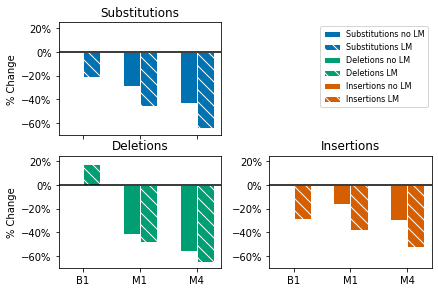}
\caption{The change in number of Levenshtein edit operations compared to our baseline B1 with best path decoding and no language model.}
\label{fig:lev_ops}
\end{figure}

To understand the benefit of fine-tuning and language model-decoding, we further analyze the types of error corrected by the respective approaches, using Levenshtein single character edit operations (as measured from gold standard to predicted utterance) as our proxy for types of errors. For this analysis, we use the L2-ARCTIC development set. An interesting finding of our analysis (see Fig. \ref{fig:lev_ops}) is that while adding language model decoding to the B1 baseline improves WER on L2-ARCTIC, it increases the number of \textit{deletion} operations, indicative of over-generation. For fine-tuned models (M1 and M4), there is reduction in error types across the board, with particular benefit to \textit{substitution} ($M4:-43\%$), and \textit{deletion} operations  ($M4:-55\%$) and mild benefit to \textit{insertion} operations  ($M4:-30\%$). For adding a language model on top of the fine-tuning, we see further reduction in the \textit{substitution} operations  ($M4_{LM}:-64\%$) and \textit{insertion} operations ($M4_{LM}:-52\%$), with mild benefit to \textit{deletion} operations ($M4_{LM}:-65\%$).

We give examples of some of these errors in Table \ref{tab:errors}, and use the B1 predictions on the dev set of L2-ARCTIC to explore them in more detail. When looking at cases of single Levenshtein edit operations we notice the following patterns: Out of the $18$ \textit{deletion} operations, $15$ are spelling errors, one is a pluralization error, one is a tense error, and one is a spacing error. For the $18$ \textit{substitution} operations all are indicative of spelling errors. Of the $11$ \textit{insertion} operations $7$ are spacing errors and $3$ are spelling errors, and one is a pluralization error. Using this rough analysis, it appears that while both fine-tuning and language model decoding substantially improve spelling of the model, language model decoding is more effective at ensuring proper spacing of words.

\begin{table*}[!t]
\small    
\caption{\label{tab:errors} 
    Examples of transcription output of different categories of edit operation. We use a shorthand to indicate the applied method, fine-tuning: $+FT$, language model decoding: $+LM$, or lack thereof ($-$). Errors capitalized.
}\begin{center}
\begin{tabular}{c|c|l}

\hline
 Edit Type &
 \multicolumn{1}{l|}{\textbf{Model}} & \multicolumn{1}{c}{\textbf{Model output}} \\
\hline
\multirow{4}{*}{Substitution} &
Ref & the portuguese boy crawled nearer and nearer\\
\hline
&  $-FT -LM$ &  the portuguese boy CROWLED nearer and nearer\\
& $-FT +LM$ &  the portuguese boy crawled nearer and nearer\\
& $+FT -LM$ &  the portuguese boy CROWLED nearer and nearer\\
& $+FT +LM$ &  the portuguese boy crawled nearer and nearer\\

\hline
\multirow{4}{*}{Insertion} &
Ref & tomorrow it will be strong enough for you to stand upon\\
\hline
 & $-FT -LM$ &  TO MORROW it will be strong enough for you to stand upon\\
& $-FT +LM$ &  TO MORROW it will be strong enough for you to stand upon\\
& $+FT -LM$ &  tomorrow it will be strong enough for you to stand upon\\
& $+FT +LM$ &  tomorrow it will be strong enough for you to stand upon\\
\hline
\multirow{4}{*}{Deletion} &
Ref & there are no kiddies and half grown youths among them\\
\hline
 & $-FT -LM$ &  there are no kiddies and half grown YOUTH among them\\
& $-FT +LM$ &  there are no kiddies and half grown YOUTH among them\\
& $+FT -LM$ &  there are no kiddies and half grown YOUTH among them\\
& $+FT +LM$ &  there are no kiddies and half grown youths among them\\
\hline
\hline
\end{tabular}
\end{center}
\end{table*}

\section{Conclusion}\label{sec:con}
We demonstrated the potential of developing accent-independent and accent-dependent models that improve non-native speech recognition simply by fine-tuning the pre-trained wav2vec 2.0 model on a small amount of labeled data. Both our multi- and single-accent models improve performance on L2 English speakers. However, each accent benefits differently: results of the multi-accent, zero-shot experiments suggest that transfer learning on accent is possible and single-accent models improve the most for the target L2 accents. Comparing the benefit from using a language model in decoding the ASR outputs with simply fine-tuning the models, we find that both these methods yield comparable improvements. We also find that the combination of the two methods greatly closes the gap between L1 and L2 ASR. We summarize our findings as follows:
\begin{itemize}
    \item Fine-tuning either on single accents (most effective) or groups of accents (more generalizeable) significantly improves L2 ASR performance. This is possible through large reductions in calculated substitution and deletion operations.
    \item Fine-tuning on domain-specific L1 accents is counterproductive to L2 ASR.
    \item Language model decoding is useful for L2 ASR, even for models with strong language model-free performance on L1 speech, and is particularly good at reducing calculated substitution and insertion operations. 
\end{itemize}

When looking at future research directions, it is important to stress the need the field has for benchmark L2 ASR datasets. Datasets proposed by~\citet{wang2020laix} is one of the few L2 ASR datasets collected with this intention in mind (although these only cover L1 Chinese speakers). Without datasets to cover a wide variety of L2 English accents, relying on accent embeddings~\cite{viglino2019end,turan2020achieving,jain2018improved} and multi-task learning might be a vital addition to L2 ASR work if the goal is wide-accent coverage. While our fine-tuning of \textit{Wav2vec 2.0-ST} did not seem to keep the the model's language model-free performance, there may be other directions to take to try to maintain it. These can include looking more closely at fine-tuning techniques, such as Layer Norm and Attention fine-tuning~\cite{li2020multilingual} or Adapter fine-tuning~\cite{houlsby2019parameter}. These might be better at preserving this internal language model, as they freeze more of the original model weights. Finally, smaller ASR model size and federated learning--although early results have noted difficulties in applying it to ASR~\cite{yu2021federated}, effort is underway to lower the training cost and improve accuracy~\cite{guliani2021training}--might bring about the potential for ASR individualization to be targeted at the level of ideolects, with people able to use ASR model's tailored to their personal accent profile. 

\begin{acknowledgement}
We would like to thank Mia Li, Jeremy Zhang, and Haejin Cho for having contributed to an initial phase of this work.
\end{acknowledgement}

\bibliographystyle{spbasic.bst}
\bibliography{l2}
\end{document}